%% file: paper24.tex
\documentclass[10pt,twocolumn,letterpaper]{article}

\usepackage{wacv}
\usepackage{times}
\usepackage{epsfig}
\usepackage{graphicx}
\usepackage{amsmath}
\usepackage{amssymb}

\usepackage{booktabs}

\def\wacvPaperID{24} 

\wacvfinalcopy 

\ifwacvfinal
\def\assignedStartPage{1} 
\fi

\ifwacvfinal
\usepackage[breaklinks=true,bookmarks=false]{hyperref}
\else
\usepackage[pagebackref=true,breaklinks=true,colorlinks,bookmarks=false]{hyperref}
\fi
\hypersetup{
    colorlinks=true,
    linkcolor=blue,
    filecolor=magenta,      
    urlcolor=magenta,
}
\ifwacvfinal
\else
\fi

\begin{document}

\title{Oriented Object Detection in Aerial Images with Box Boundary-Aware Vectors}

\author{Jingru Yi, Pengxiang Wu, Bo Liu, Qiaoying Huang, Hui Qu, Dimitris Metaxas\\
Department of Computer Science\\
Rutgers University\\
Piscataway, NJ 08854, USA\\
{\tt\small jy486@cs.rutgers.edu}
}

\maketitle

\begin{abstract}
    Oriented object detection in aerial images is a challenging task as the objects in aerial images are displayed in arbitrary directions and are usually densely packed. Current oriented object detection methods mainly rely on two-stage anchor-based detectors. However, the anchor-based detectors typically suffer from a severe imbalance issue between the positive and negative anchor boxes. To address this issue, in this work we extend the horizontal keypoint-based object detector to the oriented object detection task. In particular, we first detect the center keypoints of the objects, based on which we then regress the box boundary-aware vectors (BBAVectors) to capture the oriented bounding boxes. The box boundary-aware vectors are distributed in the four quadrants of a Cartesian coordinate system for all arbitrarily oriented objects. To relieve the difficulty of learning the vectors in the corner cases, we further classify the oriented bounding boxes into horizontal and rotational bounding boxes. In the experiment, we show that learning the box boundary-aware vectors is superior to directly predicting the width, height, and angle of an oriented bounding box, as adopted in the baseline method. Besides, the proposed method competes favorably with state-of-the-art methods. Code is available at \url{https://github.com/yijingru/BBAVectors-Oriented-Object-Detection}.
\end{abstract}

\section{Introduction}
\label{Introduction}
Object detection in aerial images serves as an essential step for numerous applications such as urban planning, traffic surveillance,  port management, and maritime rescue \cite{azimi2018towards,zhang2018toward}. The aerial images are taken from the bird's-eye view. Detecting objects in aerial images is a challenging task as the objects typically have different scales and textures, and the background is complex. Moreover, the objects are usually densely packed and displayed in arbitrary directions. Consequently, applying the horizontal bounding boxes to oriented object detection would lead to misalignment between the detected bounding boxes and the objects \cite{ding2019learning}. To deal with this problem, oriented bounding boxes are preferred for capturing objects in aerial images.

\begin{figure}[t!]
\begin{center}
   \includegraphics[width=\linewidth]{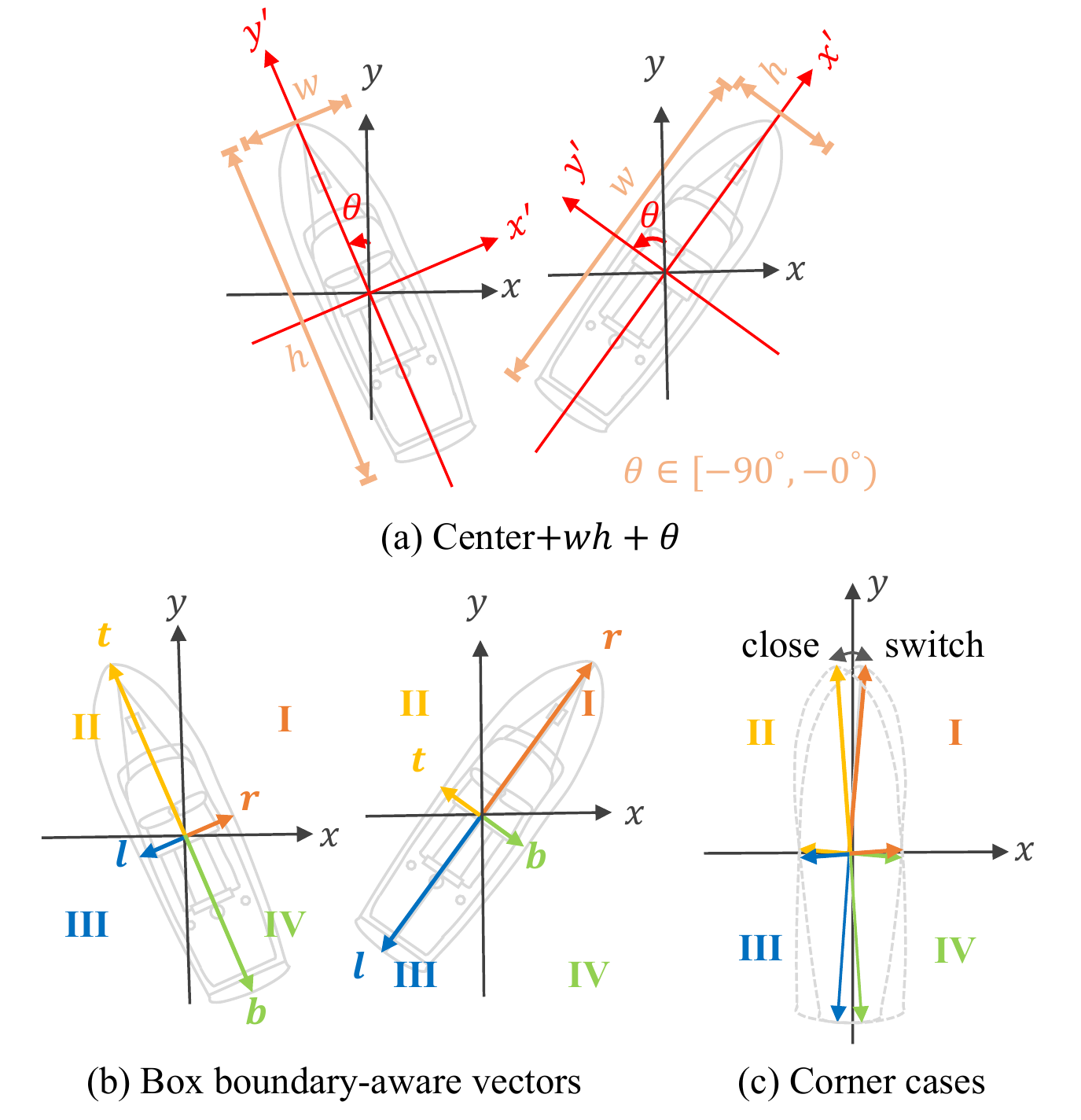}
\end{center}
\caption{Oriented bounding box (OBB)  descriptions for (a) baseline method, termed Center+$wh$+$\theta$, where $w,h,\theta$ are the width, height and angle of an OBB. Note that $w$ and $h$ of the OBBs are measured in different rotating coordinate systems for each object; (b) the proposed method, where $\textbf{t}, \textbf{r}, \textbf{b}, \textbf{l}$ are the top, right, bottom and left box boundary-aware vectors. The box boundary-aware vectors are defined in four quadrants of the Cartesian coordinate system for all the arbitrarily oriented objects; (c) illustrates the corner cases where the vectors are very close to the $xy$-axes.}
\label{fig:figure1}
\end{figure}

Current oriented object detection methods are mainly derived from the two-stage anchor-based detectors \cite{girshick2014rich,girshick2015fast,ren2015faster}. Generally, in the first stage, those detectors spread anchor boxes on the feature maps densely and then regress the offsets between the target box and the anchor box parameters in order to provide region proposals. In the second stage, the region-of-interest (ROI) features are pooled to refine the box parameters and classify the object categories. Notably, they use the center, width, height, and angle as the descriptions of an oriented bounding box. The angle is learned either in the first stage or in the second stage. For instance, R$^2$CNN \cite{jiang2017r2cnn}, Yang et al. \cite{yang2018position}, and ROI Transformer \cite{ding2019learning} regress the angle parameters from the pooled horizontal region proposal features in the second stage; similarly, R$^2$PN \cite{zhang2018toward}, R-DFPN \cite{yang2018automatic} and ICN \cite{azimi2018towards} generate oriented region proposals in the first stage. These oriented object detection methods share the same drawbacks with the anchor-based detectors. For example, the design of the anchor boxes is complicated; and the choices of aspect-ratios and the size of the anchor boxes need to be tuned carefully. Besides, the extreme imbalance between the positive and negative anchor boxes would induce slow training and sub-optimal performance \cite{duan2019centernet}. Moreover, the crop-and-regress strategies in the second stage are computationally expensive \cite{zhou2019bottom}. Recently, the keypoint-based object detectors \cite{law2018cornernet,zhou2019bottom,duan2019centernet} have been developed to overcome the disadvantages of anchor-based solutions \cite{ren2015faster,liu2016ssd,yi2019assd} in the horizontal object detection task. In particular, these methods detect the corner points of the bounding boxes and then group these points by comparing embedding distances or center distances of the points. Such strategies have demonstrated improved performance, yet with one weakness that the grouping process is time-consuming. To address this issue, Zhou's CenterNet \cite{zhou2019objects} suggests detecting the object center and regressing the width ($w$) and height ($h$) of the bounding box directly, which achieves faster speed at comparable accuracy. Intuitively, Zhou's CenterNet can be extended to the oriented object detection task by learning an additional angle $\theta$ together with $w$ and $h$ (see Fig.~\ref{fig:figure1}a). However, as the parameters $w$ and $h$ are measured in different rotating coordinate systems for each arbitrarily oriented object, jointly learning those parameters may be challenging for the model.

In this paper, we extend Zhou's CenterNet to the oriented object detection task. However, instead of regressing the $w$, $h$ and $\theta$ at the center points, we learn the box boundary-aware vectors (BBAVectors, Fig.~\ref{fig:figure1}b) to capture the rotational bounding boxes of the objects. The BBAVectors are distributed in the four quadrants of the Cartesian coordinate system.
Empirically we show that this design is superior to directly predicting the spatial parameters of bounding boxes.
In practice, we observe that in the corner cases, where the vectors are very close to the boundary of the quadrants (i.e., $xy$-axes in Fig.~\ref{fig:figure1}c), it would be difficult for the network to differentiate the vector types. To deal with this problem, we group the oriented bounding box (OBB) into two categories and handle them separately. Specifically, we have two types of boxes: horizontal bounding box (HBB) and rotational bounding box (RBB), where RBB refers to all oriented bounding boxes except the horizontal ones. We summarize our contributions as follows:
\begin{itemize}
    \item We propose the box boundary-aware vectors (BBAVectors) to describe the OBB. This strategy is simple yet effective. The BBAVectors are measured in the same Cartesian coordinate system for all the arbitrarily oriented objects. Compared to the baseline method that learns the width, height and angle of the OBBs, the BBAVectors achieve better performance.
    
    \item We extend the center keypoint-based object detector to the oriented object detection task. Our model is single-stage and anchor box free, which is fast and accurate. It achieves state-of-the-art performances on the DOTA and HRSC2016 datasets.
    
\end{itemize}

\section{Related Work}
\label{Related Work}
\subsection{Oriented Object Detection}
The horizontal object detectors, such as R-CNN \cite{ma2015hierarchical}, fast R-CNN \cite{girshick2015fast}, faster R-CNN \cite{ren2015faster}, SSD \cite{liu2016ssd}, YOLO \cite{redmon2016you}, are designed for horizontal objects detection. These methods generally use the horizontal bounding boxes (HBB) to capture the objects in natural images. Different from the horizontal object detection task, oriented object detection relies on oriented bounding boxes (OBB) to capture the arbitrarily oriented objects. Current oriented object detection methods are generally extended from the horizontal object detectors. For example, R$^2$CNN \cite{jiang2017r2cnn} uses the region proposal network (RPN) to produce the HBB of the text and combines different scales of pooled ROI features to regress the parameters of OBB. R$^2$PN \cite{zhang2018toward} incorporates the box orientation parameter into the RPN network and develops a rotated RPN network. R$^2$PN also utilizes a rotated ROI pooling to refine the box parameters.  R-DFPN \cite{yang2018automatic} employs the Feature Pyramid Network (FPN) \cite{lin2017feature} to combine multi-scale features and boost the detection performance. Based on the DFPN backbone, Yang \textit{et al}. \cite{yang2018position} further propose an adaptive ROI Align method for the second-stage box regression. RoI Transformer \cite{ding2019learning} learns the spatial transformation from the HBBs to OBBs. ICN \cite{azimi2018towards} develops an Image Cascade Network that enhances the semantic features before adopting R-DFPN. RRD \cite{liao2018rotation} uses active rotation filters to encode the rotation information. Gliding Vertex \cite{xu2020gliding} glide the vertex of the horizontal bounding boxes to capture the oriented bounding boxes. All these methods are based on anchor boxes. Overall, the anchor-based detectors first spread a large amount of anchor boxes on the feature maps densely, and then regress the offsets between the target boxes and the anchor boxes. Such anchor-based strategies suffer from the imbalance issue between positive and negative anchor boxes. The issue would lead to slow training and sub-optimal detection performances \cite{law2018cornernet}.

\begin{figure*}[t!]
\begin{center}
   \includegraphics[width=0.9\linewidth]{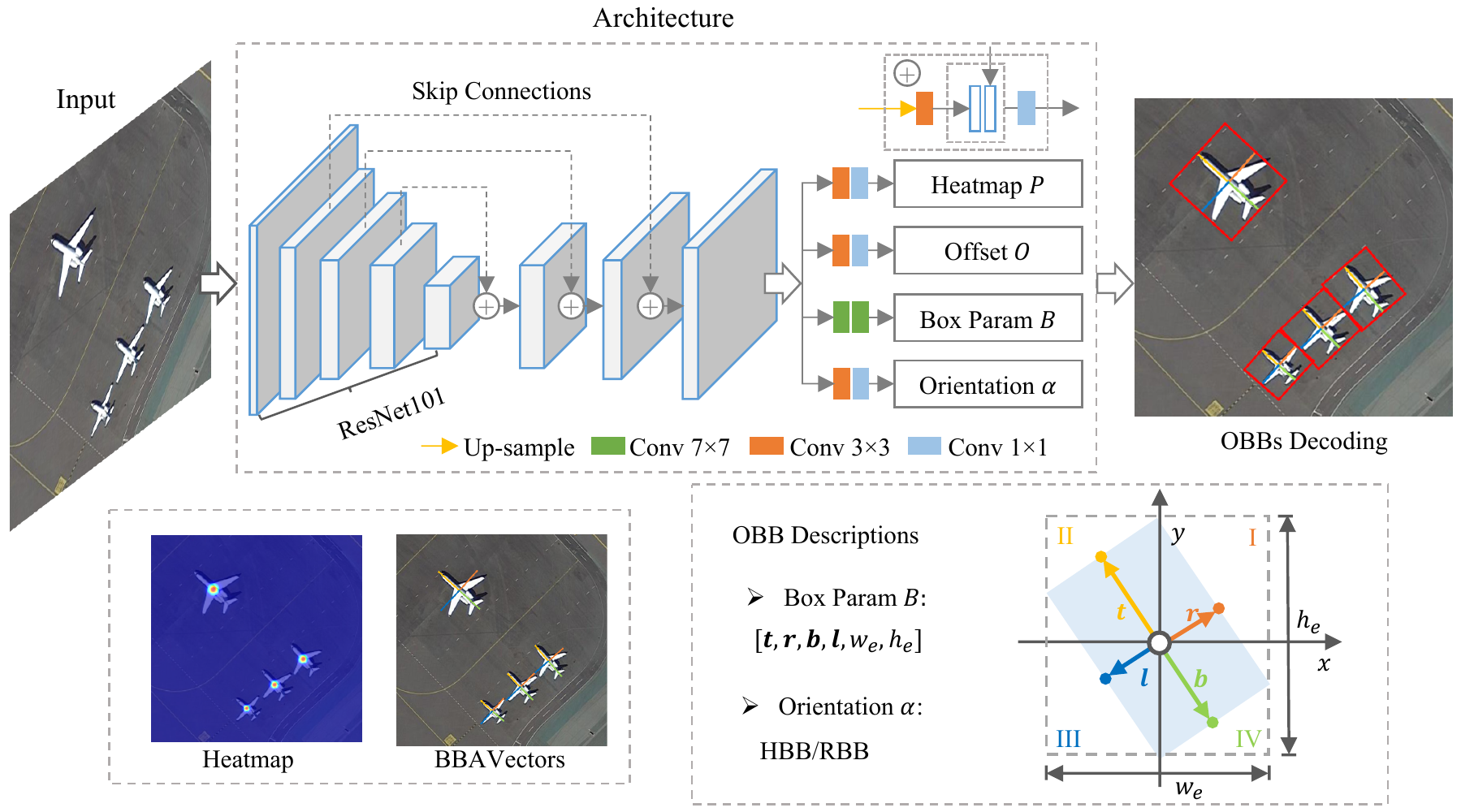}
\end{center}
\caption{The overall architecture and the oriented bounding box (OBB) descriptions of the proposed method. The input image is resized to $608\times 608$ before being fed to the network. The architecture is built on a U-shaped network. Skip connections are adopted to combine feature maps in the up-sampling process. The output of the architecture involves four maps: the heatmap $P$, offset map $O$, box parameter map $B$, and orientation map $\alpha$. The locations of the center points are inferred from the heatmap and offset map. At the center points, the box boundary-aware vectors (BBAVectors) are learned. The resolution of the output maps is $152\times 152$. HBBs refer to the horizontal bounding boxes. RBBs indicate all oriented bounding boxes except the HBBs. The symbols $\textbf{t},\textbf{r},\textbf{b}, \textbf{l}$ refer to the top, right, bottom and left vectors of BBAVectors, $w_e$ and $h_e$ are the external width and height of an OBB. The decoded OBBs are shown in red bounding boxes.}
\label{fig:figure2}
\end{figure*}

\subsection{Keypoint-Based Object Detection}
The keypoint-based object detectors \cite{law2018cornernet,zhou2019bottom,zhou2019objects,yi2019multi} capture the objects by detecting the keypoints and therefore provide anchor-free solutions. Keypoint detection is extensively employed in the face landmark detection \cite{merget2018robust} and pose estimation \cite{newell2016stacked,sun2018integral}. In the horizontal object detection task, the keypoint-based detection methods propose to detect the corner points or the center points of the objects and extract the box size information from these points. Cornernet \cite{law2018cornernet} is one of the pioneers. It captures the top-left and bottom-right corner points of the HBB using heatmaps. The corner points are grouped for each object by comparing the embedding distances of the points. Duan's CenterNet \cite{duan2019centernet} detects both corner points and center points. ExtremeNet \cite{zhou2019bottom} locates the extreme and center points of the boxes. These two methods both use the center information to group the box points. However, the post-grouping process in these methods is time-consuming. To address this problem, Zhou's CenterNet \cite{zhou2019objects} proposes to regress the width and height of the bounding box at the center point without a post-grouping process, which makes the prediction faster. The keypoint-based object detectors show advantages over the anchor-based ones in terms of speed and accuracy, yet the keypoint-based detectors are barely applied to oriented object detection task.

\paragraph{Baseline method.}  In this paper, we extend Zhou's CenterNet to the oriented object detection task. In particular, we first build a baseline method that directly regresses the width $w$ and height $h$ as well as the orientation angle $\theta$ of the bounding boxes. We term this baseline method as Center$+wh+\theta$ (see Fig.~\ref{fig:figure1}a). We compare the proposed method with Center$+wh+\theta$ to demonstrate the advantages of box boundary-aware vectors.

\section{Method}
In this section, we first describe the overall architecture of the proposed method, and then explain the output maps in detail. The output maps are gathered and decoded to generate the oriented bounding boxes of the objects.

\subsection{Architecture}
\label{sub:architecture}
The proposed network (see Fig.~\ref{fig:figure2}) is built on a U-shaped architecture \cite{ronneberger2015u}. We use the ResNet101 Conv1-5 \cite{he2016deep} as the backbone. At the top of the backbone network, we up-sample the feature maps and output a feature map that is 4 times smaller (scale $s=4$) than the input image. In the up-sampling process, we combine a deep layer with a shallow layer through skip connections to share both the high-level semantic information and low-level finer details. In particular, we first up-sample a deep layer to the same size of the shallow layer through bilinear interpolation. The up-sampled features map is refined through a $3\times3$ convolutional layer. The refined feature map is then concatenated with the shallow layer, followed by a $1\times1$ convolutional layer to refine the channel-wise features. Batch normalization and ReLU activation are used in the latent layers. Suppose an input RGB image is $I\in\mathbb{R}^{3\times H\times W}$, where $H$ and $W$ are the height and width of the image. The output feature map $X\in \mathbb{R}^{C\times \frac{H}{s}\times \frac{W}{s}}$ ($C=256$ in this paper) is then transformed into four branches: heatmap ($P\in \mathbb{R}^{K\times \frac{H}{s}\times \frac{W}{s}}$), offset ($O\in \mathbb{R}^{2\times\frac{H}{s}\times \frac{W}{s}}$), box parameter ($B\in \mathbb{R}^{10\times \frac{H}{s}\times \frac{W}{s}}$), and the orientation map ($\alpha\in \mathbb{R}^{1\times \frac{H}{s}\times \frac{W}{s}}$), where $K$ is the number of dataset categories and $s=4$ refers to the scale. The transformation is implemented with two convolutional layers with $3\times3$ kernels and $256$ channels.

\subsection{Heatmap}
\label{sub:heatmap}
Heatmap is generally utilized to localize particular keypoints in the input image, such as the joints of humans and the facial landmarks \cite{merget2018robust,newell2016stacked,sun2018integral}. In this work, we use the heatmap to detect the center points of arbitrarily oriented objects in the aerial images. Specifically, the heatmap $P\in \mathbb{R}^{K\times \frac{H}{s}\times \frac{W}{s}}$ used in this work has $K$ channels, with each corresponding to one object category. The map at each channel is passed through a sigmoid function. The predicted heatmap value at a particular center point is regarded as the confidence of the object detection.

\paragraph{Groundtruth} Suppose $\textbf{c}=(c_x, c_y)$ is the center point of an oriented bounding box, we place a 2D Gaussian  exp$({-\frac{(p_x-c_x)^2+(p_y-c_y)^2}{2\sigma^2}})$ (see Fig.~\ref{fig:figure2}) around each $\textbf{c}$ to form the groundtruth heatmap $\hat{P}\in \mathbb{R}^{K\times \frac{H}{s}\times \frac{W}{s}}$, where $\sigma$ is a box size-adaptive standard deviation \cite{zhou2019objects,law2018cornernet}, point $\hat{\mathbf{p}} = (p_x,p_y)$ indexes the pixel points on $\hat{P}$.

\paragraph{Training Loss} When training the heatmaps, only the center points $\textbf{c}$ are positive. All the other points including the points in the Gaussian bumps are negative. Directly learning the positive center points would be difficult due to the imbalance issue. To handle this problem, following the work of \cite{law2018cornernet}, we decrease the penalty for the points inside the Gaussian bumps and use the variant focal loss to train the heatmap:
\begin{equation}
    L_{h}= -\frac{1}{N}\sum_{i}
                \begin{cases}
           (1-p_{i})^\alpha\log(p_{i}) & \text{if } \hat{p}_{i}=1\\ 
               (1-\hat{p}_{i})^\beta p_{i}^\alpha \log(1-p_{i}) & \text{otherwise},\\
               \end{cases}
\end{equation}
where $\hat{p}$  and $p$ refer to the ground-truth and the predicted heatmap values, $i$ indexes the pixel locations on the feature map, $N$ is the number of objects, $\alpha$ and $\beta$ are the hyperparameters that control the contribution of each point. We choose $\alpha=2$ and $\beta=4$ empirically as in \cite{law2018cornernet}.

\subsection{Offset}
\label{sub:offsets}
In the inference stage, the peak points are extracted from the predicted heatmap $P$ as the center point locations of the objects. These center points $\textbf{c}$ are integers. However, down-scaling a point from the input image to the output heatmap generates a floating-point number. To compensate for the difference between the quantified floating center point and the integer center point, we predict an offset map $O\in \mathbb{R}^{2\times\frac{H}{s}\times \frac{W}{s}}$. Given a ground-truth center point $\Bar{\mathbf{c}} = (\Bar{c}_x,\Bar{c}_y)$ on the input image, the offset between the scaled floating center point and the quantified center point is:
\begin{equation}
    \textbf{o} = (\frac{\Bar{c}_x}{s}-\lfloor{\frac{\Bar{c}_x}{s}}\rfloor, \frac{\Bar{c}_y}{s}-\lfloor{\frac{\Bar{c}_y}{s}}\rfloor),
\end{equation}
The offset is optimized with a smooth $L_1$ loss \cite{girshick2015fast}:
\begin{equation}
    L_{o} = \frac{1}{N}\sum_{k=1}^{N}\text{Smooth}_{L_1}(\mathbf{o}_k-\hat{\mathbf{o}}_k),
\end{equation}
where $N$ is the total number of objects, $\hat{\mathbf{o}}$ refers to the ground-truth offsets, $k$ indexes the objects. The smooth $L_1$ loss can be expressed as:
\begin{equation}
    \text{Smooth}_{L_1}(x) = \begin{cases}
    0.5x^2 &\text{if }|x|<1\\
    |x|-0.5 & \text{otherwise}.
    \end{cases}
\end{equation}

\subsection{Box Parameters}
To capture the oriented bounding boxes, one natural and straightforward way is to detect the width $w$, and height $h$, and angle $\theta$ of an OBB from the center point. We term this baseline method as Center$+wh+\theta$ (see Fig.~\ref{fig:figure1}a). This method has several disadvantages. First, a small angle variation has marginal influence on the total loss in training, but it may induce a large IOU difference between the predicted box and the ground-truth box. Second, for each object, the $w$ and $h$ of its OBB are measured in an individual rotating coordinate system that has an angle $\theta$ with respect to the $y$-axis. Therefore, it is challenging for the network to jointly learn the box parameters for all the objects. In this paper, we propose to use the box boundary-aware vectors (BBAVectors, see Fig.~\ref{fig:figure1}b) to describe the OBB. The BBAVectors contain the top $\textbf{t}$, right $\textbf{r}$, bottom $\textbf{b}$ and left $\textbf{l}$ vectors from the center points of the objects. In our design, the four types of vectors are distributed in four quadrants of the Cartesian coordinate system. All the arbitrarily oriented objects share the same coordinate system, which would facilitate the transmission of mutual information and therefore improve the generalization ability of the model. We intentionally design the four vectors instead of two (i.e., $\textbf{t}$ and $\textbf{b}$, or $\textbf{r}$ and $\textbf{l}$) to enable more mutual information to be shared when some local features are obscure and weak.

The box parameters are defined as $\textbf{b}=[\textbf{t},\textbf{r},\textbf{b},\textbf{l},w_e, h_e]$, where $\textbf{t},\textbf{r},\textbf{b},\textbf{l}$ are the BBAVectors, $w_e$ and $h_e$ are the external horizontal box size of an OBB, as described in Fig.~\ref{fig:figure2}. The details of $w_e$ and $h_e$ are explained in Section \ref{sub: orientation}. Totally, the box parameter map $B\in \mathbb{R}^{10\times \frac{H}{s}\times \frac{W}{s}}$ has $10$ channels with $2\times4$ vectors and $2$ external size parameters. We also use a smooth $L_1$ loss to regress the box parameters at the center point:
\begin{equation}
    L_{b} = \frac{1}{N}\sum_{k=1}^{N}\text{Smooth}_{L_1}(\mathbf{b}_k- \hat{\mathbf{b}}_k),
\end{equation}
where $\mathbf{b}$ and $\hat{\mathbf{b}}$ are the predicted and ground-truth box parameters, respectively.

\subsection{Orientation}
\label{sub: orientation}
In practice, we observe that the detection would fail in situations where the objects nearly align with $xy$-axes (see Fig.~\ref{fig:figure4}b). The reason would be that at the boundary of the quadrant, the types of the vectors are difficult to be differentiated. We term this problem as corner cases (see Fig.~\ref{fig:figure1}c). To address this issue, in this work we group OBBs into two categories and process them separately. In particular, the two types of boxes are HBB and RBB, where RBB involves all the rotation bounding boxes except the horizontal ones. The benefit of such a classification strategy is that we transform the corner cases into the horizontal ones, which can be dealt with easily. When the network encounters the corner cases, the orientation category and the external size ($w_e$ and $h_e$ in Fig.~\ref{fig:figure2}) can help the network to capture the accurate OBB. The additional external size parameters also enrich the descriptions of an OBB. 

We define the orientation map as $\alpha\in \mathbb{R}^{1\times \frac{H}{s}\times \frac{W}{s}}$. The output map is finally processed by a sigmoid function. To create the ground-truth of the orientation class $\hat{\alpha}$, we define:
\begin{equation}
    \hat{\alpha}=\begin{cases}
    1 \text{ (RBB)} & \text{IOU(OBB, HBB)}<0.95\\
    0 \text{ (HBB)}& \text{otherwise},
    \end{cases}
\end{equation}
where IOU is the intersection-over-union between the oriented bounding box (OBB) and the horizontal bounding box (HBB). The orientation class is trained with the binary cross-entropy loss:
\begin{equation}
    L_{\alpha} = -\frac{1}{N}\sum_i^N(\hat{\alpha}_i\log(\alpha_i)+(1-\hat{\alpha}_i)\log(1-\alpha_i)),
\end{equation}
where $\alpha$ and $\hat{\alpha}$ are the predicted and the ground-truth orientation classes, respectively.

\input{tables/table1.tex}

\section{Experiments}
\subsection{Datasets}
We evaluate our method on two public aerial image datasets: DOTA \cite{xia2018dota} and HRSC2016 \cite{icpram17}.

\paragraph{DOTA.} We use DOTA-v1.0  \cite{xia2018dota} dataset for the oriented object detection. It contains 2,806 aerial images with a wide variety of scales, orientations, and shapes of objects. These images are collected from different sensors and platforms. The image resolution ranges from $800\times800$ to $4000\times4000$. The fully-annotated images contain 188,282 instances. The DOTA-V1.0 has 15 categories: Plane, Baseball Diamond (BD),    Bridge,    Ground Track Field (GTF), Small Vehicle (SV), Large Vehicle    (LV), Ship, Tennis Court (TC), Basketball Court (BC), Storage Tank (ST), Soccer-Ball Field (SBF), Roundabout (RA), Harbor, Swimming Pool (SP) and Helicopter (HC). The DOTA images involve the crowd and small objects in a large image. For accurate detection, we use the same algorithm as ROI Transformer \cite{ding2019learning} to crop the original images into patches. In particular, the images are cropped into $600\times600$ patches with a stride of 100.  The input images have two scales $0.5$ and $1$.  The trainval set and testing set contain 69,337 and 35,777 images after the cropping, respectively. The trainval set refers to both training and validation sets \cite{ding2019learning}. Following the previous works \cite{azimi2018towards,ding2019learning}, we train the network on the trainval set and test on the testing set. The detection results of cropped images are merged as the final results. Non-maximum-suppression (NMS) with a 0.1 IOU threshold is applied to the final detection results to discard repetitive detection. The testing dataset is evaluated through the online server.

\noindent\paragraph{HRSC2016.} Ship detection in aerial images is important for port management, cargo transportation, and maritime rescue \cite{yang2018automatic}. The HRSC2016 \cite{icpram17} is a ship dataset collected from Google Earth, which contains 1,061 images with ships in various appearances. The image sizes range from $300\times300$ to $1500\times900$. The dataset has 436 training images, $181$ validation images, and $444$ testing images. We train the network on the training set and use the validation set to stop the training when the loss on the validation set no longer decreases. The detection performance of the proposed method is reported on the testing set.

\subsection{Implementation Details}
We resize the input images to $608\times 608$ in the training and testing stage, giving an output resolution of $152\times 152$. We implement our method with PyTorch. The backbone weights are pre-trained on the ImageNet dataset. The other weights are initialized under the default settings of PyTorch. We adopt the standard data augmentations to the images in the training process, which involve random flipping and random cropping within scale range $[0.9,1.1]$. We use Adam \cite{kingma2014adam} with an initial learning rate of $1.25\times 10^{-4}$ to optimize the total loss $L=L_h+L_o+L_b+L_{\alpha}$. The network is trained with a batch size of 20 on four NVIDIA GTX 1080 Ti GPUs. We train the network for about 40 epochs on the DOTA dataset and 100 epochs on the HRSC2016 dataset. We additionally report an experiment with a larger batch size 48 on 4 NVIDIA Quadro RTX 6000 GPUs, we mark the results with symbol $*$ in Table \ref{Table:Table1}. The speed of the proposed network is measured on a single NVIDIA TITAN X GPU on the HRSC2016 dataset.

\subsection{Testing Details}
To extract the center points, we apply an NMS on the output heatmaps through a $3\times 3$ max-pooling layer. We pick the top 500 center points from the heatmaps and take out the offsets ($\textbf{o}$), box parameters ($\textbf{b}$), and orientation class ($\alpha$) at each center point ($\textbf{c}$). The heatmap value is used as the detection confidence score. We first adjust the center points by adding the offsets $\tilde{\mathbf{c}} =\mathbf{c} + \textbf{o}$. Next, we rescale the obtained integer center points on the heatmaps by $\bar{\mathbf{c}} = s\tilde{\mathbf{c}}, s=4$. We obtain the RBB when $\alpha>0.5$, and we get the HBB otherwise. We use the top-left ($\textbf{tl}$), top-right ($\textbf{tr}$), bottom-right ($\textbf{br}$) and bottom-left ($\textbf{bl}$) points of the bounding box as the final decoded points. Specifically, for a center point $\bar{\mathbf{c}}$, the decoded RBB points are obtained from:
\begin{align}
    \textbf{tl} &= (\textbf{t}+\textbf{l})+\bar{\mathbf{c}}, \quad
    \textbf{tr} = (\textbf{t}+\textbf{r})+\bar{\mathbf{c}}\nonumber\\
    \textbf{br} &= (\textbf{b}+\textbf{r})+\bar{\mathbf{c}},
    \quad
    \textbf{bl} = (\textbf{b}+\textbf{l})+\bar{\mathbf{c}}.
\end{align}
For a HBB, the points are:
\begin{align}
    \textbf{tl} &= (\bar{c}_x-w_e/2, \bar{c}_y-h_e/2),
    ~\textbf{tr} = (\bar{c}_x+w_e/2, \bar{c}_y-h_e/2)\nonumber\\
    \textbf{br} &= (\bar{c}_x+w_e/2, \bar{c}_y+h_e/2),
    ~\textbf{bl} = (\bar{c}_x-w_e/2, \bar{c}_y+h_e/2).
\end{align} 
We gather the RBBs and HBBs as the final detection results.

\subsection{Comparison with the State-of-the-arts}
We compare the performance of the proposed method with the state-of-the-art algorithms on the DOTA and HRSC2016 datasets. To study the impact of orientation classification, we define two versions of the the proposed method: BBAVectors$+r$ and BBAVectors$+rh$. BBAVectors$+r$ only learns the box boundary-aware vectors to detect OBB, which contains box parameters $b=[\textbf{t}, \textbf{r}, \textbf{b}, \textbf{l}]$. BBAVectors$+rh$ additionally learns the orientation class $\alpha$ and the external size parameters ($w_e$ and $w_h$).

\begin{figure*}[tbh!]
\begin{center}
   \includegraphics[width=0.92\linewidth]{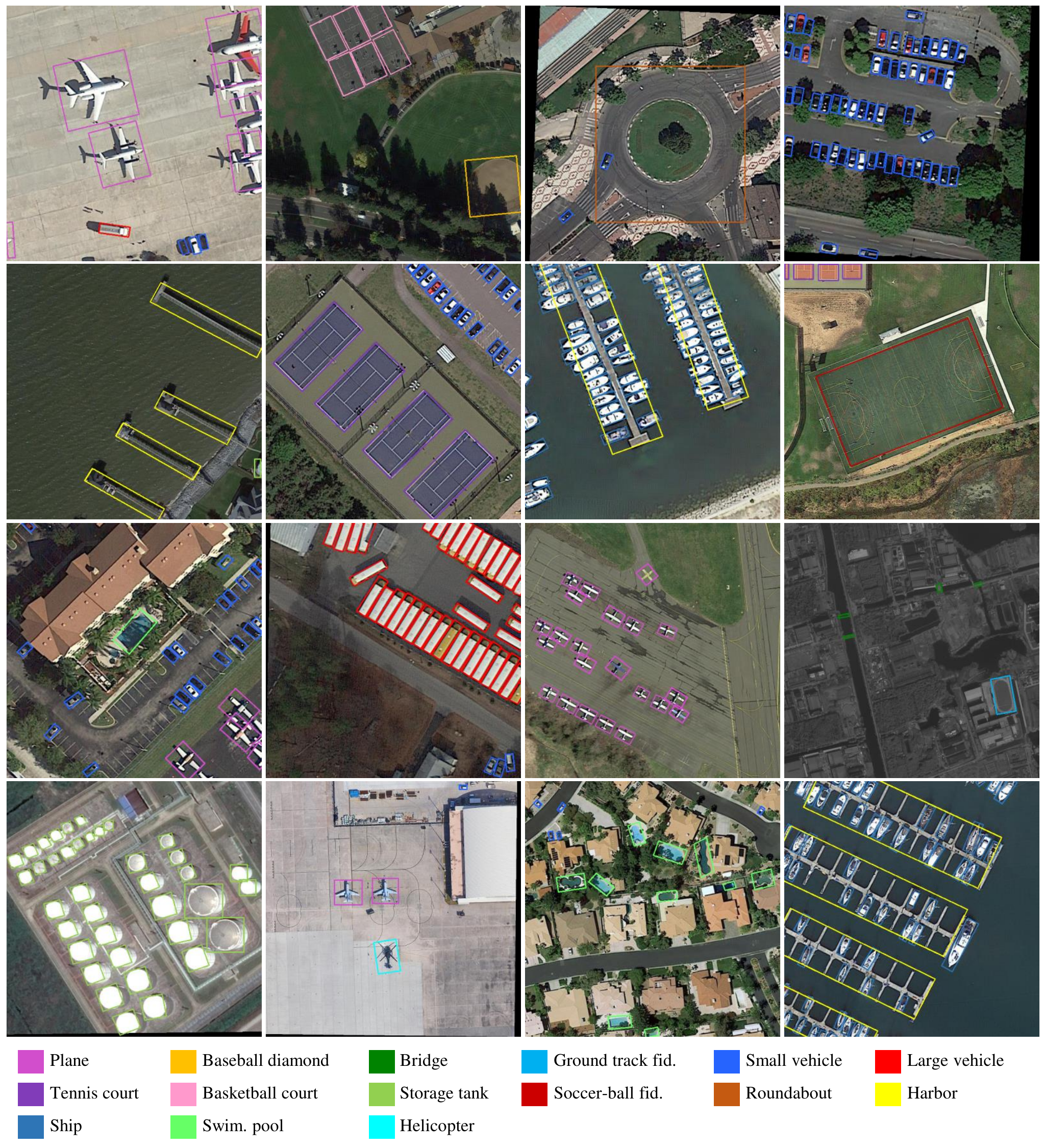}
\end{center}
\caption{Visualization of the detection results of BBAVectors+$rh$ on DOTA dataset.}
\label{fig:figure3}
\end{figure*}

\paragraph{DOTA.} The detection results on the DOTA dataset are illustrated in Table~\ref{Table:Table1}. YOLOv2 \cite{redmon2017yolo9000} and FR-O \cite{xia2018dota} are trained on HBB \cite{xia2018dota} and their performances are comparably lower than the other methods. Notably, although the one-stage detector YOLOv2 runs faster, its accuracy is lower than the two-stage anchor-based detectors. R-DFPN \cite{yang2018automatic} learns the angle parameter from faster R-CNN \cite{ren2015faster} and improves performance from 54.13\% to 57.94\%. R$^2$CNN \cite{jiang2017r2cnn} pools multiple sizes of region proposals at the output of RPN and improves the accuracy from 57.94\% to 60.67\%. Yang \textit{et al}. \cite{yang2018position} use the adaptive ROI align to extract objects and achieve 1.62\% improvement over 60.67\%. ICN \cite{azimi2018towards} adopts the Image Cascaded Network to enrich features before R-DFPN and boost the performance from 62.29\% to 68.16\%. ROI Transformer  \cite{ding2019learning} transfers the horizontal ROIs to oriented ROIs by learning the spatial transformation, raising the accuracy from 68.16\% to 69.56\%. Different from these methods, the proposed method offers a new concept of oriented object detection, a keypoint-based detection method with box boundary-aware vectors. As shown in Table \ref{Table:Table1}, Without orientation classification, the BBAVectors$+r$ improves 2.05\% over 69.59\% of ROI Transformer+FPN \cite{lin2017feature} and 3.87\% over 67.74\% of ROI Transformer without FPN. As the proposed method is single-stage, this result demonstrates the detection advantages of the keypoint-based method over the anchor-based method. With the orientation classification and additional external OBB size parameters ($w_e$ and $h_e$), the proposed BBAVectors$+rh$ achieves 72.32\% mAP, which exceeds ROI Transformer+FPN by 2.76\%. Besides, BBAVectors$+rh$ runs faster than ROI Transformer (see Table~\ref{Table:Table2}). With a larger training batch size, the BBAVectors$+rh^*$ achieves about 3 points higher than BBAVectors$+rh$. The visualization of the detection results of BBAVectors$+rh$ on DOTA dataset is illustrated in Fig.~\ref{fig:figure3}. The background in the aerial images is complicated and the objects are arbitrarily oriented with different sizes and scales. However, the proposed method is robust to capture the objects even for the tiny and crowd small vehicles.

\paragraph{HRSC2016.} The performance comparison results between the proposed method and the state-of-the-arts on HRSC2016 dataset is illustrated in Table~\ref{Table:Table2}. The R$^2$PN \cite{zhang2018toward} learns the rotated region proposals based on VGG16 backbone, achieving 79.6\% AP. RRD \cite{liao2018rotation} adopts activate rotating filters and improves the accuracy from 79.6\% to 84.3\%. ROI Transformer   \cite{ding2019learning} without FPN produces 86.2\%, while BBAVectors+$r$ achieves 88.2\%. BBAVectors$+rh$ performs slightly higher (0.4\% over 88.2\%) than BBAVectors$+r$. 
In the inference stage, the proposed method achieves 12.88 FPS on a single NVIDIA TITAN X GPU, which is 2.18x faster than ROI Transformer.

\input{tables/table2.tex}

\begin{figure}[tbh!]
\begin{center}
   \includegraphics[width=0.95\linewidth]{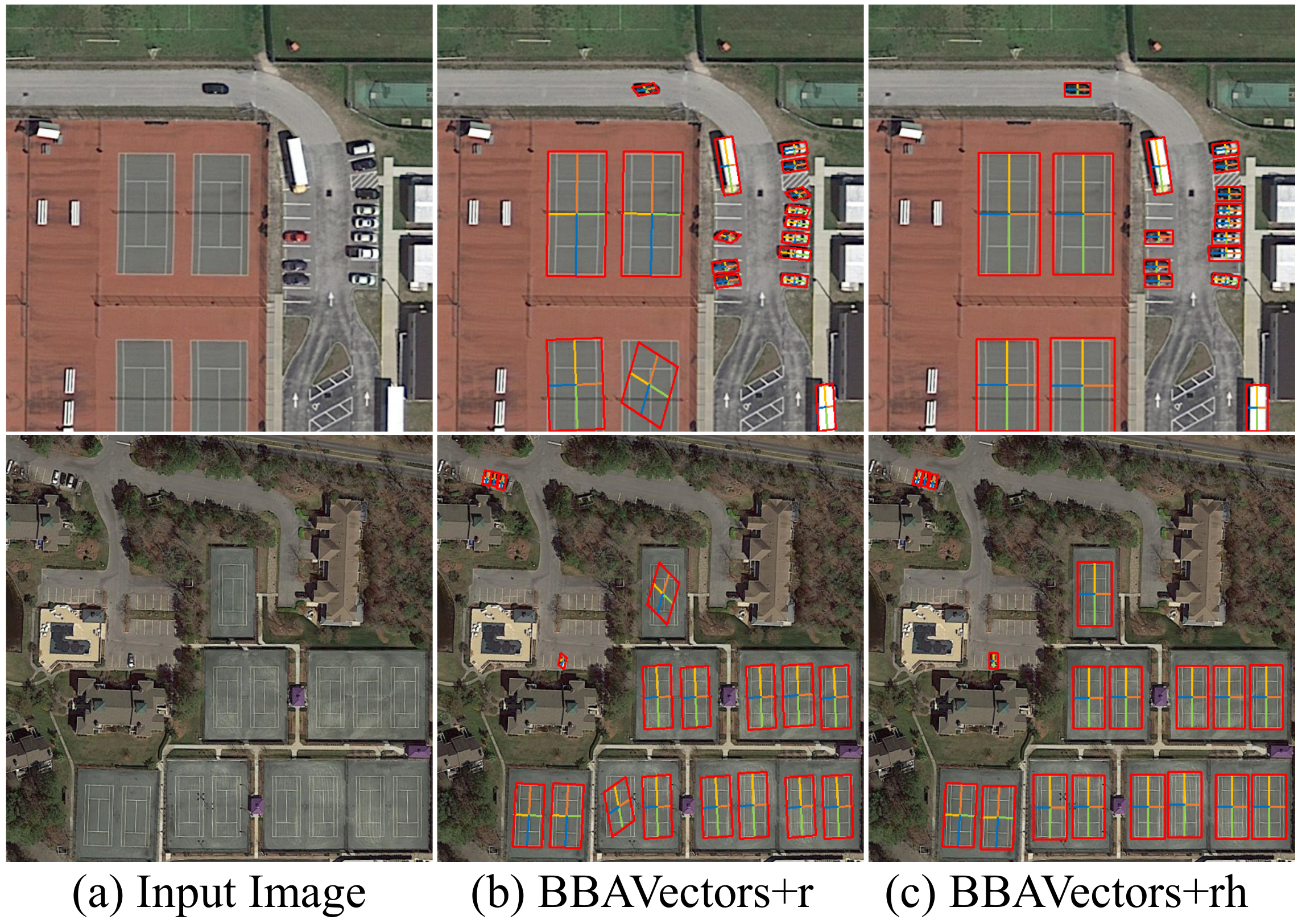}
\end{center}
\caption{Comparison of BBAVectors+$r$ and BBAVectors+$rh$.}
\label{fig:figure4}
\end{figure}

\subsection{Ablation Studies}
\label{sub:ablation studies}
We compare the performances of the proposed BBAVectors$+r$ and BBAVectors$+rh$ to study the impact of orientation classification. As we mentioned before, BBAVectors$+r$ refers to box parameters $b=[\textbf{t}, \textbf{r}, \textbf{b}, \textbf{l}]$.  BBAVectors$+rh$ corresponds to $b=[\textbf{t}, \textbf{r}, \textbf{b}, \textbf{l}, w_e, h_e, \alpha]$. BBAVectors$+r$ has $274.01$MB parameters with $25.82\times 10^{9}$ FLOPs, while BBAVectors$+rh$ has $276.28$MB parameters and $27.68\times 10^{9}$ FLOPs. 

As can be seen in Fig.~\ref{fig:figure4}, the BBAVectors$+r$ can hardly capture the bounding boxes that nearly align with the $xy$-axes. These are the corner cases as discussed above. The reason for the failure detection would be that it is difficult for the network to differentiate the type of vectors near the quadrant boundary (i.e., classification boundary). To address this problem, we separate the OBB into RBB and HBB by learning an orientation class $\alpha$ and we use the external parameters ($w_e$ and $w_h$) to describe HBB. As illustrated in Fig.~\ref{fig:figure4}, the BBAVectors$+rh$ excels in capturing the oriented bounding box at the corner cases. On the DOTA dataset (see Table~\ref{Table:Table1}), the BBAVectors$+rh$ improves 0.71\% over BBAVectors$+r$. On the HRSC2016 dataset (see Table~\ref{Table:Table2}), BBAVectors$+rh$ achieves 0.4\% improvement over BBAVectors$+r$. 

\subsection{Comparison with Baseline}
To explore the advantage of the box boundary-aware vectors, we also compare our method with the baseline Center+$wh$+$\theta$ (see Fig.~\ref{fig:figure1}a), which employs the width ($w$), height ($h$) and angle ($\theta\in(-90^\circ,-0^\circ]$) as the descriptions of OBB. Note that the baseline method shares the same architecture as the proposed method except for the output box parameter and orientation map. The training procedure is the same as the proposed method. Here we do not explicitly handle the corner cases for a fair comparison. From Table~\ref{Table:Table3}, we can see that the proposed method performs 4.82\% and 2.74\% better than Center+$wh$+$\theta$ on HRSC2016 and DOTA datasets, respectively. The results suggest that the box boundary-aware vectors are better for oriented object detection than learning the $w,h,\theta$ of OBB directly. The reason would be that the box boundary-aware vectors are learned in the same Cartesian coordinate systems, while in the baseline method, the size parameters ($w$ and $h$) of an OBB are measured in different rotating coordinate systems that have an angle $\theta$ with respect to the $y$-axis. Jointly learning those parameters would be difficult for the baseline method.

\input{tables/table3.tex}


\section{Conclusion}
In this paper, we propose a new oriented object detection method based on box boundary-aware vectors and center points detection. The proposed method is single-stage and is free of anchor boxes.  The proposed box boundary-aware vectors perform better in capturing the oriented bounding boxes than the baseline method that directly learns the width, height, and angle of the oriented bounding box. The results on the HRSC2016 and DOTA datasets demonstrate the superiority of the proposed method over the state-of-the-arts.


\input{paper24.bbl}
\end{document}

%% file: tables/table1.tex
\begin{table*}[t!]
\begin{center}
\resizebox{\textwidth}{!}{
\begin{tabular}{l|c|ccccccccccccccc}
\toprule
Method & mAP & Plane & BD & Bridge & GTF & SV & LV & Ship & TC & BC & ST & SBF & RA & Harbor & SP & HC\\
\midrule
YOLOv2 \cite{redmon2017yolo9000}& 25.49 &52.75 &24.24 &10.6 &35.5 &14.36& 2.41 &7.37 &51.79& 43.98 &31.35& 22.3& 36.68& 14.61 &22.55 &11.89\\
FR-O \cite{xia2018dota}&54.13
&79.42&77.13&17.7&64.05&35.3&38.02&37.16&89.41&69.64&59.28&50.3&52.91&47.89&47.4&46.3 \\
R-DFPN \cite{yang2018automatic}&57.94&80.92 &65.82& 33.77& 58.94 &55.77& 50.94& 54.78 &90.33& 66.34 &68.66& 48.73& 51.76& 55.10& 51.32& 35.88 \\
R$^2$CNN \cite{jiang2017r2cnn}&60.67&80.94&65.75&35.34& 67.44	&59.92	&50.91	&55.81	&90.67&	66.92&	72.39&	55.06	&52.23&	55.14&	53.35&	48.22\\
Yang \textit{et al}. \cite{yang2018position}& 62.29&81.25 &71.41 &36.53 &67.44& 61.16& 50.91& 56.60& 90.67 &68.09& 72.39 &55.06& 55.60 &62.44& 53.35& 51.47 \\
ICN \cite{azimi2018towards}&68.16& 81.36& 74.30& 47.70& 70.32& 64.89& 67.82& 69.98& 90.76& 79.06& 78.20& 53.64& \color{blue}\textbf{62.90}& \color{blue}\textbf{67.02} &64.17& 50.23\\
ROI Trans.  \cite{ding2019learning}&67.74&88.53&77.91&37.63&\color{blue}\textbf{74.08}&66.53&62.97&66.57&90.5&79.46&76.75&\color{red}\textbf{59.04}&56.73&62.54&61.29&55.56\\
ROI Trans.+FPN  \cite{ding2019learning}& 69.56& \color{red}\textbf{88.64}&78.52&43.44&\color{red}\textbf{75.92}&68.81&73.68&83.59&90.74&77.27&81.46&\color{blue}\textbf{58.39}&53.54&62.83&58.93&47.67\\
BBAVectors+$r$ & 71.61& 88.54& 76.72& 49.67& 65.22& 75.58& \color{blue}\textbf{80.28}& 87.18 & 90.62& \color{blue}\textbf{84.94}&\color{blue}\textbf{84.89}& 47.17& 60.59& 65.31& 63.91& 53.52\\
BBAVectors+$rh$&\color{blue}\textbf{72.32}& 88.35& \color{blue}\textbf{79.96}& \color{blue}\textbf{50.69}& 62.18& \color{red}\textbf{78.43}& 78.98& \color{blue}\textbf{87.94}& \color{blue}\textbf{90.85}& 83.58& 84.35& 54.13& 60.24& 65.22& \color{blue}\textbf{64.28}& \color{blue}\textbf{55.70}\\
BBAVectors+$rh^*$&\color{red}\textbf{75.36}& \color{blue}\textbf{88.63} & \color{red}\textbf{84.06} & \color{red}\textbf{52.13} & 69.56 & \color{blue}\textbf{78.26} & \color{red}\textbf{80.40} & \color{red}\textbf{88.06} & \color{red}\textbf{90.87} & \color{red}\textbf{87.23} & \color{red}\textbf{86.39} & 56.11 & \color{red}\textbf{65.62}& \color{red}\textbf{67.10} & \color{red}\textbf{72.08} & \color{red}\textbf{63.96}\\
\bottomrule
\end{tabular}
}
\end{center}
\caption{Detection results on the testing set of DOTA-v1.0. The performances are evaluated through the online server. Symbol $*$ shows the result with a larger training batch size (i.e., 48 on 4 Quadro RTX 6000 GPUs). {\color{red}Red} and {\color{blue}Blue} colors label the best and second best detection results in each column.}
\label{Table:Table1}
\end{table*}

%% file: tables/table2.tex
\begin{table}
\begin{center}
\resizebox{\linewidth}{!}{
\begin{tabular}{l|c|c|c|c|c}
\toprule
Method & Backbone & Image Size&GPU& FPS & AP \\
\midrule
R$^2$PN \cite{zhang2018toward}&VGG16&-&-&-&79.6\\
RRD \cite{liao2018rotation}&VGG16&$384\times384$&-&-&84.3\\
ROI Trans. \cite{ding2019learning}&ResNet101&$512\times 800$&TITAN X&5.9&86.2\\
BBAVectors+$r$&ResNet101&$608\times608$&TITAN X&12.88&88.2\\
BBAVectors+$rh$&ResNet101&$608\times608$&TITAN X&11.69&\textbf{88.6}\\
\bottomrule
\end{tabular}}
\end{center}
\caption{Detection results on the testing dataset of HRSC2016. The speed of the proposed method is measured on a single NVIDIA TITAN X GPU.}
\label{Table:Table2}
\end{table}

%% file: tables/table3.tex
\begin{table}
\begin{center}
\resizebox{0.95\linewidth}{!}{
\begin{tabular}{l|c|c|c}
\toprule
Method & Dataset & Backbone& mAP \\
\midrule
Center+$wh$+$\theta$ & HRSC2016 & ResNet101 & 83.40\\
BBAVectors+$r$ & HRSC2016 & ResNet101& \textbf{88.22}\\
Center+$wh$+$\theta$ & DOTA & ResNet101& 68.87\\
BBAVectors+$r$ & DOTA  & ResNet101& \textbf{71.61}\\
\bottomrule
\end{tabular}
}
\end{center}
\caption{Comparison between baseline method Center+$wh$+$\theta$ and the proposed method BAVectors$+r$. }
\label{Table:Table3}
\end{table}
